\documentclass[11pt,a4paper]{article}
\usepackage{times,latexsym}
\usepackage{url}
\usepackage[T1]{fontenc}

\usepackage[acceptedWithA]{tacl2018v2}

\usepackage{subcaption}

\usepackage{xspace,mfirstuc,tabulary}

\newif\iftaclinstructions
\taclinstructionsfalse 
\iftaclinstructions
\renewcommand{\confidential}{}
\renewcommand{\anonsubtext}{(No author info supplied here, for consistency with
TACL-submission anonymization requirements)}
\newcommand{\instr}
\fi

\iftaclpubformat

\else

\fi

\usepackage{color,soul}
\usepackage{amsmath}
\DeclareMathOperator*{\argmax}{arg\,max}

\usepackage{dsfont}
\usepackage{todonotes}
\usepackage{arydshln}
\usepackage{enumitem}
\usepackage{booktabs}
\usepackage{cleveref}

\newcommand{\nullout}[1]{}

\usepackage{microtype}

\title{Iterative Paraphrastic Augmentation with Discriminative Span Alignment}

 \author{
Ryan Culkin ~~ J. Edward Hu ~~ Elias Stengel-Eskin ~~ Guanghui Qin ~~ Benjamin Van Durme \\
  Johns Hopkins University\\
  {\tt \{rculkin, edward.hu, elias, qin\}@jhu.edu} \\
  {\tt vandurme@cs.jhu.edu}
}

\date{}

\begin{document}
\maketitle
\begin{abstract}
We introduce a novel paraphrastic augmentation strategy based on sentence-level lexically constrained paraphrasing and discriminative span alignment. Our approach allows for the large-scale expansion of existing resources, or the rapid creation of new resources from a small, manually-produced seed corpus. We illustrate our framework on the Berkeley FrameNet Project, a large-scale language understanding effort spanning more than two decades of human labor. Based on roughly four days of collecting training data for the alignment model and approximately one day of parallel compute, we automatically generate 495,300 unique $(\texttt{Frame}, \texttt{Trigger})$ combinations annotated in context, a roughly 50x expansion atop FrameNet v1.7.
\end{abstract}

\begin{figure*}
    \centering
    \includegraphics[scale=.41]{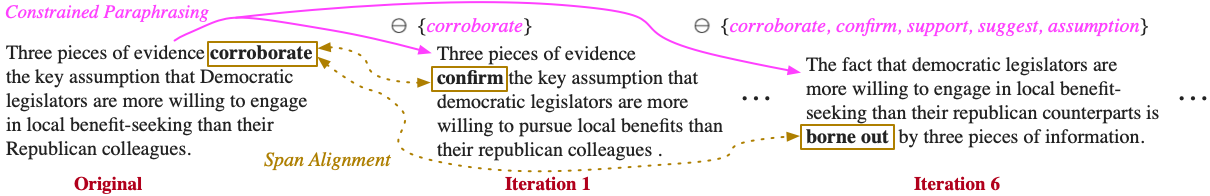}
    \caption{Framework for iterative paraphrastic augmentation illustrated on an actual system output.  The original, manually-annotated sentence contains a tag over the word ``corroborate".  In Iteration 1, the sentence is paraphrased using a lexically constrained decoder with a negative constraint on ``corroborate" and all associated inflectional forms, guaranteeing that it will not appear in the output.  Then, a span alignment model is used to obtain a link between 
    ``corroborate" in the original sentence and ``confirm" in the paraphrased sentence.  All inflectional forms of ``confirm" are then unioned with the set of negative constraints and the process repeats for a predetermined number of iterations.}
    \label{fig:iteration-example}
\end{figure*}

\section{Introduction}

\textit{Data augmentation} is the process of automatically increasing the size of a dataset with the goal of improving performance on a task of interest.  It has been applied in many areas of machine learning including computer vision \cite{shorten.c.2019} and
speech recognition \cite{ragni.a.2014,ko.t.2015}.

In this paper, we focus on \textit{paraphrastic augmentation}, a technique to automatically expand text-based datasets both in their overall size and in their degree of lexical and syntactic diversity, via the use of a paraphrase model.  Broadly speaking, a paraphrase model outputs a sentence $S'$ given an input sentence $S$ such that $\texttt{meaning}(S) \approx \texttt{meaning}(S')$ and $S \neq S'$.  Prior work has demonstrated the efficacy of paraphrastically augmented datasets on a variety of sentence-level tasks, including machine translation, natural language inference, and intent classification (e.g. \cite{ribeiro-etal-2018-semantically,hu.e.2019,kumar.a.2019}).  
Here we focus on augmenting data for span labeling problems, where we are concerned with balancing the joint objectives of finding different ways to express meaning at the level of a word or phrase while ensuring the paraphrase is sensitive to the context of the surrounding sentence.

Often in paraphrastic augmentation an input sentence is rewritten one or more times, with the assumption the output(s) are meaning preserving.  For example, in sentiment analysis, data consists of $(\texttt{Sentence}_i, \texttt{Label}_i)$ pairs, where each $\texttt{Label}_i$ is in $\{0, 1\}$, indicating negative or positive sentiment.  To augment this kind of dataset, we can  paraphrase each $\texttt{Sentence}_i$ with a model $f$ and thereby produce an additional $(f(\texttt{Sentence}_i), \texttt{Label}_i)$ pair, doubling the size of the dataset. 

In many language understanding tasks however, data contains \emph{span} labels of the form: $(\texttt{Sentence}_i, \{(\texttt{start}_{i,1},\texttt{end}_{i,1},\texttt{type}_{i,1}),...\})$, where the latter element is a set of tuples indicating each label's location (as a contiguous subsequence of the input tokens) and type.  
Although a paraphrase is expected to have the same meaning as the sentence from which it was generated, words and phrases are usually added, removed, or reordered.  For a given annotated sentence, while we expect the same label \emph{types} to still apply to a paraphrase, the \emph{location} (start and end) will likely shift.
To address this issue, we introduce a new model for span-based discriminative sentence alignment.  Given an input sentence $S$, a paraphrase $f(S)$, and a span of tokens in $S$ representing a label location, the alignment model finds a semantically equivalent span in $f(S)$.  We present the architectural details of this model, a dataset for span alignment, and corresponding results in \S \ref{alignment-section}.

A second problem is that most paraphrase models offer no control over specific words or phrases that are included in or excluded from the final output.  Text-based data augmentation typically aims to increase lexical diversity, so it would be useful to force each tagged text span in the input to be rewritten in the paraphrase, ideally as a synonymous or semantically similar phrase via \textit{lexically constrained decoding} (\S \ref{sentential-section}).

Finally, we describe in \S \ref{iterative-section} a framework that utilizes constrained paraphrasing and alignment in conjunction, iteratively, to augment datasets for span labeling problems.  A schematic is given in Figure \ref{fig:iteration-example}. We present the results of applying this framework to FrameNet in \S \ref{application-section}, including a new dataset with 495,300 unique $(\texttt{Frame}, \texttt{Trigger})$ pairs annotated in context.\footnote{\url{http://nlp.jhu.edu/parabank}}

\section{Background}

\paragraph{Monolingual Paraphrasing}
Coinciding with the improvement of machine translation, several works have explored sentential paraphrasing through back-translations~\cite{paranet,paranmt}. One such model~\cite{paranmt} was used for sentence canonicalization, although its further usefulness was hindered by the lack of control over the paraphrasing process. ~\newcite{parabank} introduced constrained decoding~\cite{fast_lexical} to sentential paraphrasing, enabling lexical control over the paraphrases.

\paragraph{Automatic Lexicon Expansion}
As an alternative to manual labor, past work has sought to automatically build on existing semantic resources. \newcite{semantic_taxonomy} used hypernym predictions and coordinate term classifiers to add 10,000 new WordNet entries with high precision.
FrameNet+~\cite{framenet-fast-paraphrastic-tripling-of-framenet} tripled the size of FrameNet by substituting words from PPDB~\cite{ganitkevitch2013ppdb}, a collection of primarily word-level paraphrases obtained via bilingual pivoting.  The paraphrases lack context, so e.g., ``quite" might be listed as a paraphrase of ``especially", without any means to determine when one might not be an appropriate substitute. While the expansion itself involved little cost, the lexicalized nature of their procedure failed to capture word senses in context and resulted in many false positives, requiring costly manual evaluation of every sentence.   In contrast, we seek to mitigate false positives and enhance lexical and syntactic diversity by using a context-aware paraphrase model.

\paragraph{Paraphrasing for Structured Prediction}
Structured prediction finds a mapping between a surface form and some aspect of its underlying structure. Natural language allows for surface forms that express the same meaning -- paraphrases -- which makes learning this mapping nontrivial.  \newcite{berant14} leveraged unstructured Q\&A data by learning a paraphrasing model which maps a new query to existing ones with known structures. More relevant to our work, \newcite{wang15} built a semantic parser from a small seed lexicon by generating canonical utterances from a domain-general grammar and then manually collecting paraphrases of these utterances through crowd-sourcing. A semantic parser is then trained on the paraphrases to produce the underlying structures that generated them. Our work is distinct in that we \textit{automatically} expand our seed lexicon, collecting human judgments on a small subset of outputs in order to assess quality.  Moreover, we introduce a general framework for augmenting data for span labeling, while \newcite{wang15} focused on parsing.

\paragraph{Monolingual Span Alignment} 
\newcite{yao.x.2013a} introduce a discriminatively-trained CRF model for monolingual word alignment, expanded to span alignment by \newcite{yao.x.2013b}. \newcite{ouyang.j.2019} introduced a pointer-network-based phrase-level aligner for paraphrase alignment which obtains high recall on several tasks. Syntactic chunking is used to build a candidate set of phrases in both source and paraphrase sequences, which the model is then tasked with aligning. Their model is applied to an open alignment task, where more than one phrase in the source and paraphrase should be aligned, differing from the setting described in \S \ref{alignment-section}.

\paragraph{The Berkeley FrameNet Project}

\begin{figure}
    \centering
    \includegraphics[scale=0.65]{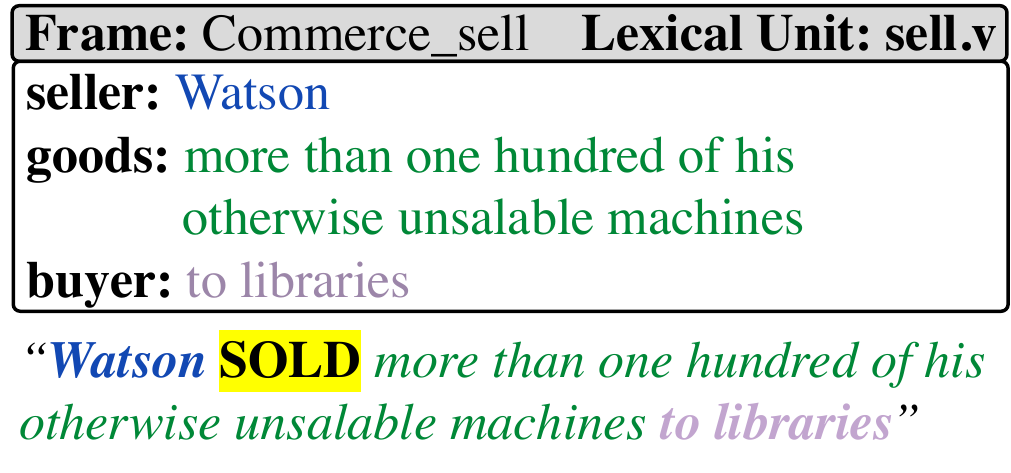}
    \caption{An example annotation from FrameNet. The trigger, ``sold", an instance of the \texttt{sell.v} lexical unit, evokes the \texttt{Commerce\_sell} frame.  The participating entities, or \textit{frame elements}, are represented as colored text.}
    \label{fig:fn-example}
\end{figure}

FrameNet \cite{Baker2007} is the application of frame-semantic theory \cite{FrameSemantics} to real-world data. 
Organizationally, each \textit{frame} contains a description of a concept, a list of entities participating in the frame (\textit{frame elements}), and a list of \textit{lexical units}, which are the semantically similar words that evoke, or \textit{trigger}, the given concept. Figure \ref{fig:fn-example} illustrates a sentence labeled under the FrameNet protocol.  As of FrameNet v1.7, the resource contains roughly 1,200 frames, 8,500 annotated lexical units, and 200,000 annotations.

FrameNet has been applied to a variety of NLP tasks, including semantic role labeling \cite{Gildea:2002:ALS:643092.643093}, question-answering \cite{shen-lapata-2007-using}, information extraction \cite{ruppenhofer-rehbein-2012-semantic},
and recognizing textual entailment \cite{Burchardt2006ApproachingTE}.
As an entirely manually-created resource, FrameNet's utility is limited by the size of its lexical inventory and number of annotations \cite{shen-lapata-2007-using,framenet-fast-paraphrastic-tripling-of-framenet}; an ideal candidate for augmentation.

\section{Lexically Constrained Paraphrasing}
\label{sentential-section}
Sentential paraphrasing is a sequence generation problem where the goal is to find an output sequence conveying similar semantics to the input sequence while also ensuring that the two sequences are lexically or syntactically distinct. Prior work has approached this problem with sequence-to-sequence neural networks~\cite{paranmt,hu.e.2019}, where an encoder embeds the input sequence into a fixed-dimensional space and a decoder produces a sequence autoregressively. Often, the decoder uses beam search to explore the output space more efficiently.
\textit{Lexically constrained} decoding allows one to dynamically include or exclude token sequences from the output via user-supplied positive or negative constraints.
When combined with paraphrasing, it can boost external NLP task performance via data augmentation~\cite{hu.e.2019}.
Our work employs negative constraints, which exclude certain token sequences from the output. This is achieved by setting the likelihood of the last tokens in the sequences to zero when the preceding tokens were generated~\cite{hu.e.2019}.

We recreated the rewriter described in the prior work by using a paraphrase corpus~\cite{hu.e.2019-CoNLL} that offers richer lexical diversity. We followed the model architecture described in ~\citeauthor{hu.e.2019} with a few minor changes: 1) we use SentencePiece~\cite{sentencepiece} unigrams instead of tokenization, following ~\citeauthor{hu.e.2019-CoNLL}; 2) we do not not use source factors, as SentencePiece unigrams are case-sensitive. These changes allow us to rewrite raw text without tokenization. 
\footnote{Our rewriter is a Transformer with a 6-layer encoder, a 4-layer decoder, 8 attention heads, an embedding size of 512, and a feed-forward size of 2048. It is trained until convergence on all sentence pairs from ~\newcite{hu.e.2019-CoNLL} with the reference sentence as target.}

\section{Alignment Models}
We present a BERT-based model \cite{DBLP:journals/corr/abs-1810-04805} to align spans of text between paraphrastic sentence pairs.  The model is trained and evaluated on a new dataset released alongside this paper, consisting of 36,417 labeled sentence pairs.

\label{alignment-section}
\subsection{Word Alignment Baselines} We compare our span alignment model (\S \ref{span-alignment-model-section}) with two word-level alignment baselines: FastAlign \cite{dyer.c.2013} and DiscAlign \cite{stengel-eskin.e.2019}. The former is a fast implementation of IBM Model 2 \cite{brown.p.1993} which decomposes the conditional probability of a target sequence given a source sequence into a lexical model and an alignment model. FastAlign is an asymmetric model, meaning that it must be run in both directions (source to paraphrase and paraphrase to source) and then these alignments must be combined using some heuristic\textemdash we use the \emph{grow-diag-final-and} heuristic. A FastAlign model was run over the concatenation of the test data, the train data, and paraphrased FrameNet data to obtain the final test alignments. 

DiscAlign is a discriminatively-trained neural alignment model which uses the matrix product of contextualized encodings of the source and paraphrase word sequences to directly model the probability of an alignment given the source and paraphrase sequences. Unlike FastAlign, which is trained on bitext alone, DiscAlign is pre-trained on bitext and fine-tuned on gold-standard alignments. For this task, a DiscAlign model was pre-trained with 141 million sentences of ParaBank data~\cite{parabank} and finetuned on a 713 sentence subset of the Edinburgh++ corpus \cite{cohn.t.2008}. Both DiscAlign and FastAlign have been successfully used for cross-lingual word alignment, with DiscAlign outperforming FastAlign on Arabic-English and Chinese-English alignment by a large margin \cite{stengel-eskin.e.2019}.

\subsection{Span Alignment Model}

 \paragraph{Architecture}
 \label{span-alignment-model-section}
  Our model takes as input two tokenized English-language sentences $\mathbf{S}$ (\textit{source}, with $n$ tokens) and $\mathbf{S'}$ (\textit{reference}, with $m$ tokens), where $\mathbf{S'}$ is a paraphrase of $\mathbf{S}$. The model also takes as input a span $\mathbf{s}$ in $\mathbf{S}$: a contiguous subsequence of tokens with length between 1 and $n$, initially represented as a tuple of $(\texttt{start}, \texttt{end})$ offsets into the source-side token sequence.  Given this, the model predicts a span $\mathbf{\hat s} \in \{(i,j) | 1 \leq i \leq j \leq m\}$, representing the best alignment between $\mathbf{s}$ and the $O(n^2)$ possible candidate spans\footnote{The model only explicitly scores the $O(n)$ reference-side spans whose length is within $k$ of the source-side span. Remaining spans are implicitly assigned zero probability. } in $\mathbf{S'}$.

In the forward pass, we embed $\mathbf{S}$ and $\mathbf{S'}$ using a pre-trained 12-layer BERT-Base model with frozen parameters, obtaining a hidden vector $t_i \in \mathds{R}^{768}$ for each of the $(m+n+3)$ input tokens.
$\mathbf{S}$ and $\mathbf{S'}$ are embedded at the same time, i.e. as
$\texttt{[CLS]} \; \mathbf{S} \; \texttt{[SEP]} \; \mathbf{S'} \; \texttt{[SEP]}$
, following the Microsoft Research Paraphrase Corpus \cite{dolan-brockett-2005-automatically} paraphrase classification experiments of \newcite{DBLP:journals/corr/abs-1810-04805}.

We then obtain a fixed-size representation $\mathcal{S} \in \mathds{R}^{768}$ of the source-side span by mean-pooling the hidden states corresponding to the token positions from the start offset $\mathbf{s}_1$ to the end offset $\mathbf{s}_2$.
In the same way, we compute span representations $\mathcal{C}_i$ for each of the $O(n)$ reference-side candidate answer spans whose length\footnote{In our experiments we used $k=5$; this was the lowest value that guaranteed the gold-standard reference-side span would be considered as a possible candidate 100\% of the time in the training set.} is within $k$ of the length of the source-side span $\mathbf{s}$. 
For each span pair representation $(\mathcal{S}, \mathcal{C}_i)$ we create an aggregate  $\mathcal{V}_i \in \mathds{R}^{1540}$ by concatenating  three vectors:

\begin{itemize}[noitemsep,nolistsep]
    \item Element-wise difference (Df): $\mathcal{S} - \mathcal{C}_i$
    \item Element-wise maxima (Mx): $\texttt{max}(\mathcal{S}, \mathcal{C}_i)$
    \item Positional cues (Cue): start index and length per span\footnote{This vector contains four elements: the start index and length corresponding to the $\mathcal{S}$ representation, and the start index and length corresponding to the $\mathcal{C}_i$ representation.}
\end{itemize}

Intuitively, if the element-wise difference of the two span representations is close to the zero vector, the spans are likely close in meaning.  Concatenating element-wise maxima to the representation worked best empirically, suggesting that extreme values may contain information not present in other parts of the representation. Since word spans in the source likely start in a similar position and are of a similar length as compared to corresponding word spans in the reference, the positional cues provide a useful signal.  The aggregate vector $\mathcal{V}_i$ is fed into a simple feedforward neural network $f$, consisting of one layer with 770 hidden units, PReLU activations, batchnorm, and a sigmoid output layer.

We use binary cross entropy loss with \textit{soft} labels: rather than each $\mathcal{C}_i$ candidate span being labeled as 1 or 0 depending on whether it is the gold-standard span or not, we assign labels according to the function, $2^{-d(\mathcal{S}, \mathcal{C}_i)}$,
where $d$ measures the absolute difference of the start and end offsets between two spans, $d(a, b) = |a_1 - b_1| +  |a_2 - b_2|$.
In this way, the gold span is given a label of 1, candidate spans that are close to the gold-standard span are given partial credit, and partial credit exponentially decreases towards 0 as the distance between the candidate span and gold-standard span increases.  In Tables \ref{tab:soft_span_res} and \ref{tab:hard_span_res} this modification is referred to as ``soft binary cross entropy", or SBCE.

\begin{figure}
    \centering
    
    \includegraphics[width=.97\columnwidth]{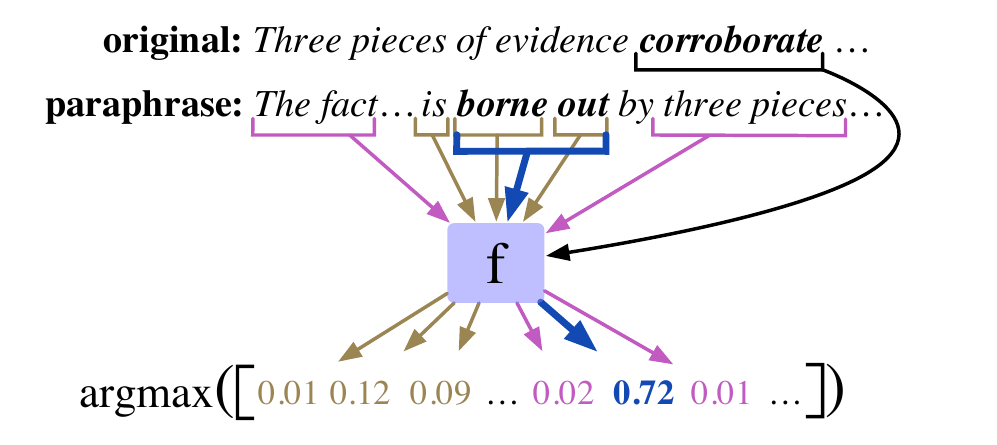}
    \caption{Span alignment inference.  A BERT-based representation of the input span ``corroborate" is passed to a neural network \textit{f} that scores the input span  against each possible candidate span.}
    \label{fig:model-example}
\end{figure}

At inference time, we choose the span corresponding to the aggregate representation $\mathcal{V}_i$ that is assigned the highest score by the neural network $f$, i.e. $\mathbf{\hat{s}} = \argmax _i f(\mathcal{V}_i) $.  A diagram illustrating the inference procedure is given in Figure \ref{fig:model-example}.

\paragraph{Data}
To train and evaluate our model we crowdsourced a span-alignment dataset consisting of 36,417 labeled sentence pairs, which we release to the community. Each instance in the dataset consists of a sentence, a span in the sentence, an automatic paraphrase, and a span in the automatic paraphrase, where the two spans have been manually aligned.  The source text was taken from FrameNet, which already has span annotations, so we fixed these spans and asked annotators to identify the corresponding spans in each automatic paraphrase.  Annotators were given the option to decide that no semantically equivalent phrase was present, which occurred roughly 9\% of the time. Of the cases where annotators did select a span, they chose the same span approximately 88\% of the time.  The text content under the original sentence spans was diverse, with roughly 10k unique phrases; approximately 4 alignments per phrase.

\subsection{Results}
\begin{table}[]\small
    \centering
    \begin{tabular}{l|c  c  r  }
    \toprule
    {\bf Method} & {\bf P} & {\bf R} & {\bf F${_1}$} \\
    \midrule
        DiscAlign & 34.11 & 39.69 &  36.69 \\
        FastAlign & 78.64 & 72.13 & 75.25\\
        \midrule
        Df+Mx+Cue+SBCE & \textbf{96.75} & \textbf{88.24} & \textbf{92.30} \\
    \bottomrule
    
    \end{tabular}
    \caption{Soft-match span F${_1}$ on dev.}
    \label{tab:soft_span_res}
\end{table}

Since the baseline aligners are word-level, and our model is span-level, in order to have a fair comparison we evaluate on span F${_1}$ (Table \ref{tab:soft_span_res}), computing the overlap between the reference span in the paraphrase and the predicted span. 
Predicted spans are obtained from word-level alignments by following the alignments of each word in the reference span to the paraphrase, and taking the maximal span covered by those alignments. 
The span F${_1}$ metric allows partial credit to be assigned to the model in cases where the predicted span and reference span do not match exactly. 

We also evaluate spans with exact matching (Table \ref{tab:hard_span_res}), where credit is only assigned if the predicted span matches the gold span exactly.
Table \ref{tab:soft_span_res} shows that when evaluated on span overlap, our model significantly outperforms both baselines.

\begin{table}[]\small
    \centering
    \begin{tabular}{l|c  c  r  }
    \toprule
    {\bf Method} & {\bf P} & {\bf R} & {\bf F${_1}$} \\
    \midrule
        DiscAlign &  (29.82) & (29.82) & 29.82 \\
        FastAlign &  (71.02) & (71.02) & 71.02 \\
        \hline
        Cue & 10.39 &	9.77 &	10.07 \\
        Mx & 80.65 &	77.92 &	79.26 \\
        Df & 87.31 &	85.42 &	86.36 \\
        Mx+Cue & 87.5 &	86.49 &	86.99 \\
        Df+Cue & 88.74 &	86.96 &	87.84 \\
        Df+Mx & \textbf{89.27} &	87.29 &	88.27 \\
        Df+Mx+Cue & 89.15 &	88.19 &	88.67 \\
        Df+Mx+Cue+SBCE & 89.14 &	\textbf{88.99} &	\textbf{89.06} \\
    \bottomrule
    \end{tabular}
    \caption{Exact-match span F${_1}$ on dev.  (Disc,Fast)Align are both word-alignment models, where ours were trained for span-alignment.}
    \label{tab:hard_span_res}
\end{table}

Table \ref{tab:hard_span_res} shows that these results generalize to the more difficult exact match setting. While all models experience a drop in performance, our model continues to outperform both baselines. Because no prediction threshold was used in the baselines (unlike in our model) the values for precision and recall are equal for the baselines but can differ slightly for our model, as the addition of a threshold allows the model to incur a false negative without predicting a false positive.

\subsection{Discussion}
Tables \ref{tab:soft_span_res} and \ref{tab:hard_span_res} reflect the strength of our model for span alignment.
Because our model is trained to choose spans by design, the probability of an exact match is higher a priori since its task is more constrained: rather than choosing the words of a span independently, it chooses them as a set, with limits on the difference in length between the source and target spans.
This is reflected both in the better performance of our model on the exact match as well as the soft matching evaluation (where an exact match counts as perfect precision and perfect recall, greatly boosting scores).

The last two rows of Table \ref{tab:hard_span_res} illustrate that SBCE boosts recall while keeping precision virtually intact; our intuition is that this training regime gives the model more confidence at inference time when scoring spans which appear similar to, but slightly different from the assumed correct answer, where those spans were then ultimately correct.

To determine whether our model was simply memorizing information associated with each lexical unit, we ran an experiment where all source-side spans in the test set were guaranteed to not have been observed at training time\footnote{In our main experiments, (original sentence, trigger, paraphrase, alignment) combinations are disjoint between train and test, but it is possible to observe the same trigger (with a different sentence, paraphrase, or alignment) at both train- and test-time.}.  Under this setting, we lost roughly two points of F${_1}$, suggesting that the model generalizes well to unseen words.

\section{Iterative Augmentation Procedure}
\label{iterative-section}
Our alignment model (\S \ref{alignment-section}) is paired with the lexically constrained paraphrase model (\S \ref{sentential-section}) to form an iterative procedure for augmenting data of the form: $(\texttt{Sentence}_i, \{(\texttt{start}_{i,1},\texttt{end}_{i,1},\texttt{type}_{i,1}),...\})$.  The process consists of three steps: constraint expansion, paraphrasing, and aligning.  In constraint expansion, we negatively constrain on a text span of interest, including its upper/lowercase counterparts and morphological variants using the \texttt{pattern} software package~\cite{smedt2012pattern}.  By applying negative constraints, the paraphrase model is forced to generate a semantically equivalent sentence with a different surface form of the labeled text, thereby increasing the size of the lexicon.  In the alignment stage, we score representations of each candidate span in the paraphrase together with the representation of the original text span, selecting the one with the highest score under the model. Using the newly obtained aligned phrase as the input to constraint expansion, we repeat the process for a predetermined number of iterations.

To encourage the model to produce as many new words as possible, we perform \textit{frame-wise constraint unioning}: taking the union of all the constraint sets from sentences that originated in the same frame, and then using that as the constraint set for those sentences in the next iteration. This prevents the same lexical unit from being used by rewrites of different example sentences in the same frame.

\section{Experiments}
\label{application-section}

Our approach lends itself to two scenarios: in \S \ref{reconstructing-subsection} we are concerned with building a semantic resource from scratch, whereas in \S \ref{expanding-subsection} we are concerned with expanding a pre-existing resource.  We demonstrate the usefulness of our approach on downstream tasks in \S \ref{sec:ie}, where we apply our generated paraphrastic dataset to the task of Frame Identification.  Following \newcite{framenet-fast-paraphrastic-tripling-of-framenet}, we consider FrameNet as an illustrative resource motivating augmentation. In all experiments we treat each system output (paraphrase and alignment) as evoking the same frame as the original FrameNet input.

\subsection{Building FrameNet (nearly) from Scratch}
\label{reconstructing-subsection}
To simulate constructing a resource using iterative paraphrastic augmentation we consider what FrameNet would have looked like in its earliest stages of development\footnote{The decision to select our seeds based on frame creation date -- in contrast to some other sub-selection strategy -- was informed by discussions with FrameNet creators (personal communication).}.   Using each object's ``created date" attribute, we ablate out all but the 20 earliest-added frames, the three earliest-added lexical units per frame, and the three earliest-added annotations per lexical unit, for a total of at most\footnote{In practice we were left with slightly fewer (171), as we removed sentences that were observed by the alignment model at training time, and some lexical units contained less than three annotations.} 180 annotations in our seed corpus.

We then ran 10 iterations of augmentation with a beam size of 30 for the paraphrase model. For each input, we ran the alignment model on each of the top-20 beam elements and chose the beam element with the highest score under the alignment model.  At the end of each iteration, constraints were unioned frame-wise.  This resulted in 1710 paraphrased and aligned sentences\footnote{I.e., 171 sentences rewritten 10 times each.}, and 1316 unique $(\texttt{Frame}, \texttt{LexicalUnit})$ combinations.  Some generated words lemmatized to the same form, causing the number of lexical units to be less than the number of sentences.

\paragraph{Automatic Evaluation}
Prior to ablation, the 20 frames in the seed corpus contained a total of 360 lexical units, of which 60 were chosen to remain in the seed. We treat the set of 300 unobserved lexical units as gold standard and compute precision and recall of the lexical units contained within the 1710-sentence system output.  Lexical units were only considered correct if they were in the correct frame; comparisons were made between $(\texttt{Frame}, \texttt{LexicalUnit})$ combinations.

Our system produced 128 true positives, 1188 false positives, and 172 false negatives, yielding a precision of 9.7\% and recall of 42.7\%.  If we include the 60 lexical units from the seed corpus, recall increases to 52.7\% of the total 360.  Although we recover over half of the lexical units, there are many false positives.  Upon manual inspection, we found that many of the words predicted by the framework were valid, yet absent from FrameNet, motivating us to develop a more comprehensive method of evaluation.

\begin{figure*}
    \centering
    \includegraphics[scale=.58]{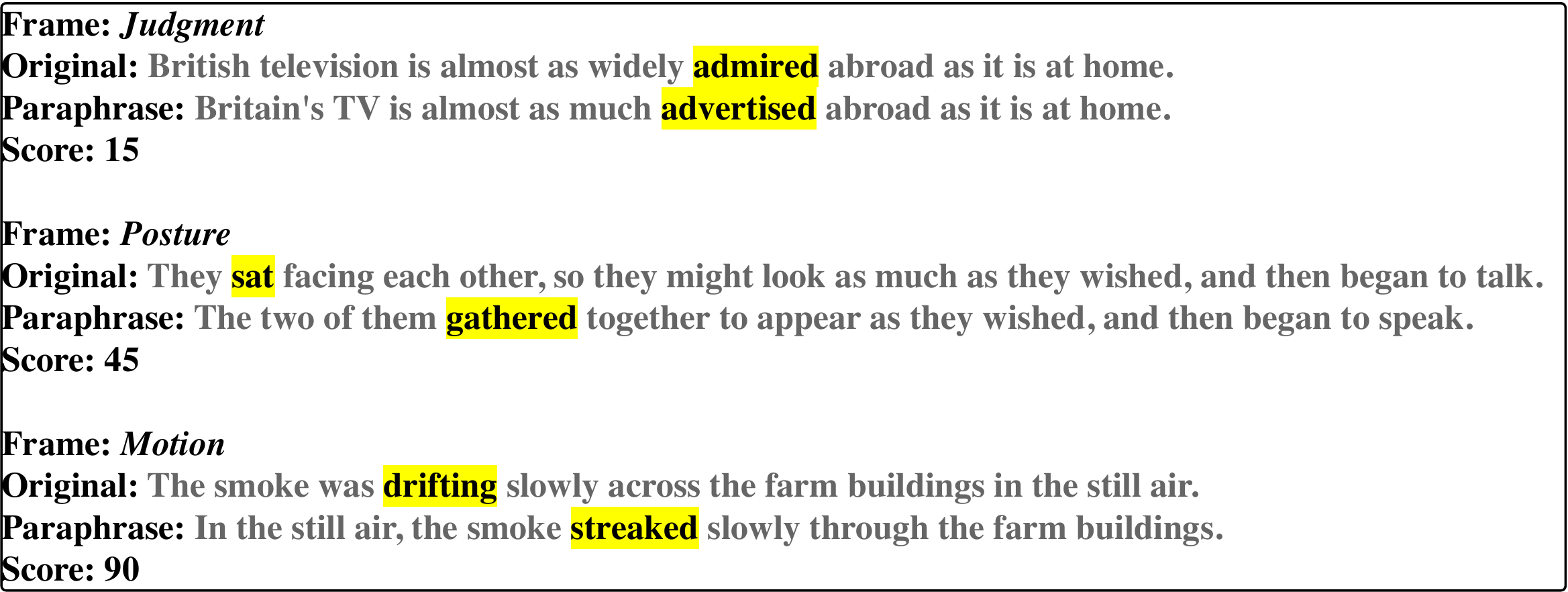}
    \caption{Sample of actual system outputs and associated manually judged scores.  Annotators did not have access to the original sentence when assigning scores, but they are provided here to illustrate the way in which the paraphrase and alignment models function.  In the first example, the paraphrase model makes a mistake; in the second, the sentence is roughly synonymous but borderline out-of-frame; in the third, both the paraphrase and alignment are high-quality.}
    \label{fig:score-example}
\end{figure*}

\paragraph{Manual Evaluation}
We conducted a 3x-redundant manual evaluation of the 1710 system outputs using trusted, locally trained annotators.  For each system output -- a paraphrase with a highlighted phrase corresponding to the span predicted by the alignment model -- we provided a description of the anticipated frame\footnote{We assume that the paraphrase transformation is label-preserving, so the anticipated frame is simply the frame of the original FrameNet sentence.} and three gold-standard example annotations\footnote{The trigger words in the example sentences were made to be disjoint with the trigger words in the candidate sentences in order to avoid biasing annotators.} to reinforce the frame definition.  Workers were then asked to rate three candidate sentences, each with a highlighted trigger phrase, on a scale of 0-100, as to how well the highlighted trigger evoked the given frame in the context of the sentence.  Unknown to annotators, of the three candidate sentences in each task, only one of them (in a random position) was an actual system output; the other two were positive or negative gold-standard sentences taken from FrameNet:

\begin{enumerate}[nolistsep,noitemsep]
    \item System output.  Frame $a$ and lexical unit $b$.
    \item Gold in-frame sentence. Frame $a$ and lexical unit $\neg b$.
    \item Gold out-of-frame adversarial example. Frame $\neg a$.
\end{enumerate}

The scores collected on gold in- and out-of-frame control sentences provide a means to ground the interpretation of scores on system outputs and also enable us to gauge overall annotator understanding of the task by scoring sentences for which we know the correct response.

Since each system output was judged by three distinct annotators, we average each triple of judgments and treat values less than 50 as a rejection (``the highlighted trigger, in the context of the sentence, does not evoke the given frame") and values greater than or equal to 50 as an acceptance.  Gold in- and out-of-frame sentences had acceptance rates of 95.26\% and 6.57\% respectively, suggesting workers possessed a relatively strong understanding of the task.  Figure \ref{fig:score-example} provides a sample of actual system outputs and associated individual (non-aggregated) scores.

\paragraph{Filtering Methods}
\label{filter-section}
We experiment with several methods of filtering system outputs, providing a trade-off between the competing goals of quality and size.  Each system output has an associated iteration number, score under the paraphrase model, and score under the alignment model; each filtering method then uses this information to select a subset of the unfiltered system outputs.  

We report the precision -- the ratio of elements in the subset that had a score over 50 -- and recall -- the number of elements in the subset with a score over 50, divided by the number of elements in the unfiltered set that also had a score over 50 -- in Table \ref{tab:manual-eval}. The upper section of Table \ref{tab:manual-eval} presents results for a variety of heuristic filtering methods, e.g. the subset of system outputs with an iteration number of three or lower, while the lower section presents results for a neural filtering model.

The neural model takes as input a system output's iteration number, score under the paraphrase model, and score under the alignment model, and produces a score between 0 and 1, where 0 represents a decision to filter an output, and 1 represents a decision to keep it.  Architecturally, the model is a feed-forward neural network with two hidden layers, 10 units per hidden layer, and a sigmoid output layer, trained to minimize binary cross entropy loss. We trained one model to favor precision by downweighting the training loss when the label was 1, and a second model to favor recall by downweighting when the label was 0.  As training data, we used the 1710 aggregated manual judgments from above (where each system output has a label of 0 or 1), plus 2988 additional judgments collected specifically for this model.  We split the data as 90\% train (4228) and 10\% test (470), and present results\footnote{Results in the upper section of Table \ref{tab:manual-eval} are reported over the 1710 system outputs from \S \ref{reconstructing-subsection}, while the results in the lower section of the table are reported over the 470-element test set.} in the lower section of Table \ref{tab:manual-eval}.

\paragraph{Discussion}
The upper section of Table \ref{tab:manual-eval} suggests that iteration number, paraphrase model score, and aligner model score each have slightly different filtering characteristics, and a simple conjunction of criteria achieves higher precision than any condition alone.  The P-Classifier, optimized to select a high-precision subset of the data, achieves higher precision than any of the heuristic methods, and higher recall than the highest-precision heuristic method. The precision of the P-classifier (95\%) is roughly the same as the human-level acceptance rate on gold in-frame sentences (95.26\%) while generating a resource that is 2.28x as large as the original.  A higher recall subset may be obtained with the R-Classifier, which retains 96.99\% of acceptable outputs with a precision of 81.19\%.

\begin{table}[h]\small
    \centering
    \begin{tabular}{l|l r r}
    \toprule
    {\bf Filtering} & {\bf P} & {\bf R} & {\bf Multiple} \\
    \midrule
        Unfiltered & 68.25 \nullout{(1167/1710)} & 100 \nullout{(1167/1167)} & 11x \\
        \hdashline
        Iter $=1$ & 90.06 \nullout{(154/171)} & 13.20 \nullout{(154/1167)} & 2x \\
        Iter $\leq 3$ & 81.29 \nullout{(417/513)} & 35.73 \nullout{417/1167} & 4x  \\
        
        PBR score $\leq 0.6$ & 90.14 \nullout{(64/71)} & 5.48 \nullout{64/1167} & 1.42x \\
        
        PBR score $\leq 0.8$ & 74.86 \nullout{(402/537)}& 34.45 \nullout{402/1167} & 4.14x  \\
        Aligner score $\geq .99$ & 85.01 \nullout{(380/447)} & 32.56 \nullout{380/1167} & 3.61x  \\
        Aligner score $\geq .95$ & 76.72 \nullout{(992/1293)} & 85.00 \nullout{992/1167} & 8.56x  \\
        Lax conjunction  & 87.73 \nullout{(243/277)} & 20.82 \nullout{243/1167} & 2.62x  \\
        Strict conjunction & 92.54 \nullout{(62/67)} & 5.31 \nullout{62/1167} & 1.39x \\
        \midrule
        P-Classifier & {\bf 95.00} \nullout{(57/60)} & 15.61 \nullout{(57/365)} & 2.28x \\
        
        R-Classifier  & 81.19 \nullout {(354/436)} & {\bf 96.99} \nullout{(354/365)} & 10.27x  \\

    \bottomrule
    \end{tabular}
    \caption{Human evaluation of system outputs across several filtering methods, with manually-judged {\bf P}recision for the subset of outputs remaining after applying the given filter, {\bf R}ecall of sentences manually judged to be acceptable, and the {\bf Multiple} (in terms of number of sentences) of the resulting dataset in relation to the original seed corpus.  {\bf Filtering} methods consider the iteration number, and scores from the paraphrase and aligner models for a given system output. The ``lax'' row applies a filter consisting of the conjunction of the criteria from rows 3, 5, and 7 (relatively lenient conditions) whereas the ``strict'' row conjoins the criteria from rows 2, 4, and 6 (which are stricter, and lead to higher precision but fewer lexical units).}

    \label{tab:manual-eval}
\end{table}

\subsection{Expanding Existing FrameNet}
\label{expanding-subsection}

In this section we report the results of applying large-scale iterative augmentation to an existing resource. As in our reconstruction experiment, we ran 10 iterations of augmentation, but with minor configuration changes to enable faster processing over the roughly 200,000 FrameNet annotations\footnote{In practice, we filtered out sentences with greater than 80 tokens due to a limitation in the paraphrase model, leaving 198,368, or 99.55\% of the original sentences.}.  The paraphrase model used a beam size of 20 and we ran the alignment model on each of the top-3 beam elements,  choosing the beam element with the highest score under the alignment model.  We did not perform frame-wise constraint unioning.

Our unfiltered dataset, which excludes the original FrameNet data, contains 1,983,680 automatically paraphrased and aligned sentences and 495,300 $(\texttt{Frame}, \texttt{Trigger})$ combinations\footnote{A $(\texttt{Frame}, \texttt{Trigger})$ combination can be thought of as an inflected surface form of a given word sense.} annotated in context. Of the 495,300 new triggers, 428,416 are unique after applying lemmatization; each lemma has 4.63 automatic in-context annotations on average. We use the filter models from \S \ref{filter-section} to select high quality and high quantity subsets of the unfiltered data; each system output in our data release has an associated score from both filter classifiers to enable post-hoc filtering.  The P-Classifier retains 138,797 sentences and 33,332 $(\texttt{Frame}, \texttt{Trigger})$ combinations, while the R-classifier retains 1,807,235 sentences and 425,050 combinations. To enable further experimentation, each sentence in our release contains a unique identifier linking it to FrameNet v1.7.

Because our data only contains alignments of triggers and not frame elements, it cannot be directly used for full FrameNet SRL. However, by additionally applying \textit{positive} constraints on frame element spans during lexically constrained decoding, an alignment link may be trivially obtained, allowing our framework to be used for full SRL.

\subsection{Using Paraphrastic Data on a Downstream Task}
\label{sec:ie}
We have demonstrated the usefulness of iterative paraphrastic augmentation for expanding lexical resources but have not shown how the resulting data is useful for downstream tasks, other than as a means to guide future lexicographical additions.  The dataset generated in \S \ref{expanding-subsection} naturally lends itself to several downstream tasks such as word sense disambiguation \cite{Das2010SEMAFOR1A} or Frame Identification, a major subtask \cite{das2010probabilistic,hermann-etal-2014-semantic} of FrameNet semantic role labeling (SRL).  In this section, we show how paraphrastic augmentation can improve Frame ID model robustness in low-resource settings.

Given an ontology in a new domain, it is often prohibitively expensive to annotate entire documents, and full-document annotation may not provide full coverage of the ontology due to the rarity of some ontological types.
A commonly-used alternative to full-document annotation is exemplar-based annotation, where several canonical examples (or "exemplars") are identified for each ontological type, ensuring at least full coverage of the ontology.
Below, we conduct experiments to show that the addition of paraphrastic data to full-document and exemplar annotations boosts Frame Identification model performance.

\paragraph{Task}
FrameNet parsing \citep{Das2014,Kshirsagar2015,Roth2015,Swayamdipta2018}, is an established task in the field of semantic parsing.
Most previous work has viewed FrameNet parsing as a semantic role labeling task, where the goal is to identify the frame and label all the arguments given a sentence with a known trigger span, but little attention has been paid to identifying trigger spans themselves \citep{Das2014}.

Given the practical importance of finding triggers, we focus on jointly identifying \emph{both} triggers and frames, rather than frames alone.

Specifically, given a sentence consisting of a sequence of words,
our task is to find all substrings\footnote{Following convention of \citet{Das2014}, we do not capture discontinuous trigger spans.
E.g., we treat \textit{there would be} as a span for the lexical unit \textit{there be.V}}
of the sentence that trigger a frame and to identify the corresponding frames.
We pose this as a span tagging problem,
with trigger spans being tagged with the associated frame and non-trigger spans tagged as \texttt{NULL}.\footnote{A few examples (0.05\% of the full-text) of FrameNet contain triggers that trigger two frames,
and we discard the second frame for simplicity.}

\paragraph{Model}
We adopt a two-pass Long Short-Term Memory (LSTM) model for the frame identification task.
We first convert the sentence $\mathbf{s} = \langle s_1, s_2, \dots, s_I \rangle$
into a sequence of embedding vectors $\langle \mathbf{e}_1^{0}, \mathbf{e}_2^{0}, \dots, \mathbf{e}_I^{0} \rangle$,
where each embedding $\mathbf{e}_i^0$ is a concatenation of
GloVe, BERT (first subtoken, fixed), character and POS embeddings \citep{Pennington2014,DBLP:journals/corr/abs-1810-04805,Alberti2019}.
Then we use a $l$-layer stacked bidirectional LSTM model \citep{Hochreiter1997} to obtain a contextual embedding for each word:
$$\langle \mathbf{e}_1^{l}, \mathbf{e}_2^{l}, \dots, \mathbf{e}_I^{l} \rangle 
= \mathrm{BiLSTM}(\langle \mathbf{e}_1^{0}, \mathbf{e}_2^{0}, \dots, \mathbf{e}_I^{0} \rangle).$$
We then apply another unidirectional LSTM model on top to get a representation for a span $\mathbf{s}_{i:j}$:
$$\mathbf{e}_{i:j} = \mathrm{LSTM}(\langle \mathbf{e}_i^{l}, \mathbf{e}_{i+1}^{l}, \dots, \mathbf{e}_j^{l}\rangle).$$
As in the alignment model,
we set a maximum span length to reduce the computation complexity from $O(I^2)$ to $O(I)$
\footnote{We empirically set the maximum span length as 3. 
The 0.24\% examples excluded by this choice are treated as false negative during evaluation.}.
A fully-connected neural network is then applied to transform the representation $\mathbf{e}_{i:j}$ into a logit vector,
which is then translated by softmax into a distribution over the label set 
comprised of frames and \texttt{NULL}.
We train with cross-entropy loss.

The FrameNet corpus provides two sets of annotated sentences: full-text and exemplars,
where the full-text contains fully annotated documents, 
but the exemplars are only annotated with one frame for every sentence.
For the full-text sentences, we treat both the trigger and non-trigger spans as training examples,
but the non-trigger spans in the exemplar and paraphrastic sentences are excluded due to the fact that they represent incomplete annotations, rather than true negative examples. 
Furthermore, 
\citet{Das2014} pointed out that some triggers are not annotated in the full-text sentences,
leading to false negative training examples.
In light of this, we apply the label smoothing trick 
\citep{Szegedy2016}
\footnote{A smoothing factor 0.2 is empirically chosen.}
on negative examples to smooth the point distribution,
resulting in a 3 F1 score improvement.

\paragraph{Experiments}
To illustrate the utility of our method in a low-resource setting,
we use a 10\% sample of the full-text sentences as our full-text dataset,
choosing the first $n_l$ lexical units by order of appearance,
and subsequently sampling $n_e$ exemplar sentences for each lexical unit.
We augment the dataset by adding the top-$n_p$ paraphrases 
(ranked by the product of paraphrase and alignment model scores) 
for each exemplar sentence.
In our experiments, we try combinations of 
$(n_l, n_e, n_p) \in \{1, 3\} \times \{1, 3\} \times \{0, 1, 4\}$,
where $n_p = 0$ means only exemplar sentences are used for training.

We use the FrameNet v1.7 release as the dataset,\footnote{We use the FrameNet support within  NLTK~\citep{Schneider2017} to process the raw data.} 
and adopt the same development and test split as proposed by \citet{Das2011},
 treating all the other documents as training examples.
We use the greedy search to find the optimal hyper-parameters
and conduct all the experiments under the same hyper-parameters.

The evaluation metric used is the frame identification F1 score,
where a frame prediction is viewed as true positive when the trigger span and frame both match exactly.

\begin{figure}[ht]
    \centering
    \includegraphics[width=1.0\linewidth]{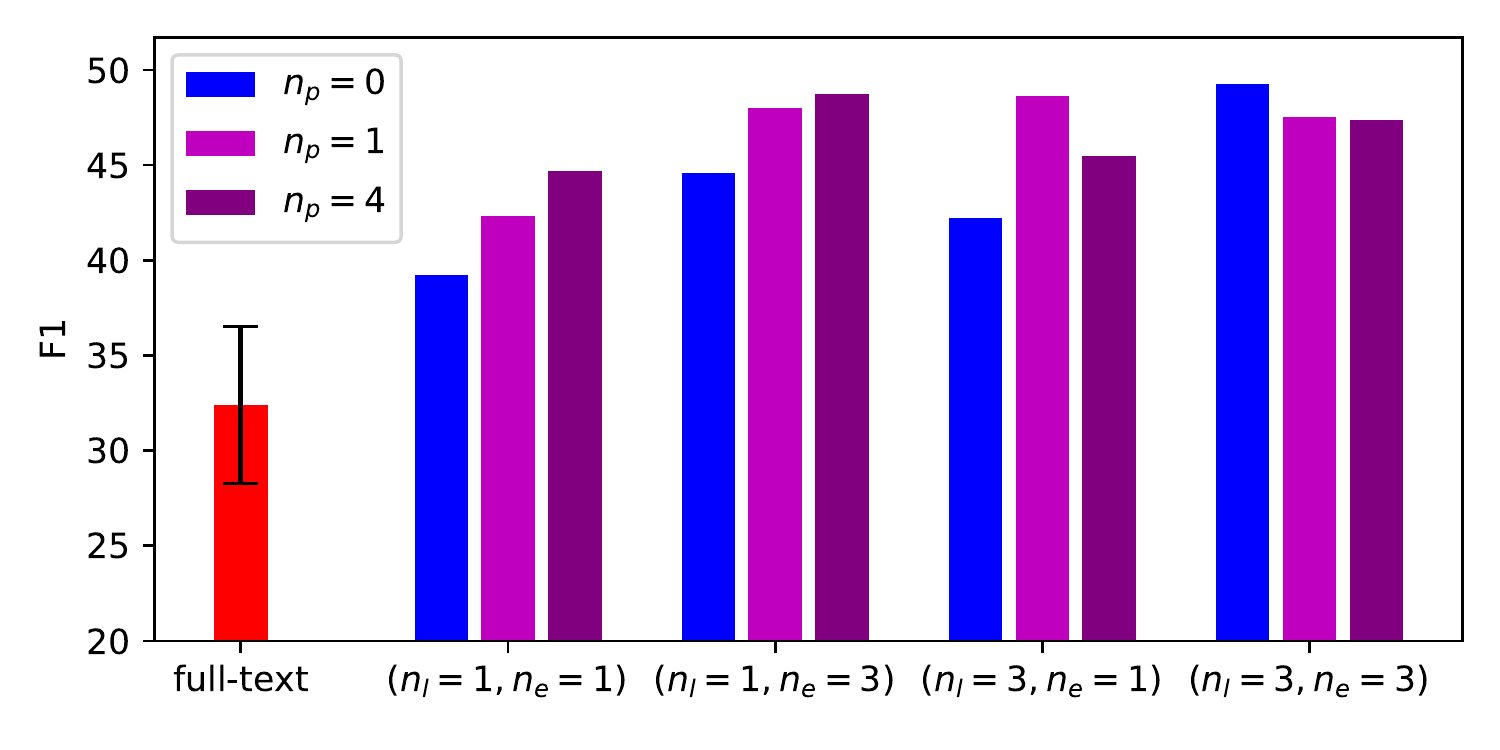}
    \caption{Frame identification results with 10\% full-text, exemplar and paraphrastic sentences.}
    \label{fig:id}
\end{figure}

\paragraph{Results and Analysis}
Results are shown in Figure \ref{fig:id},
where the leftmost bar is the result of 10\% full-text only results for reference with 5 repetitions.

We see that in both experiments increased numbers of exemplars and paraphrases improve frame identification. When we only have one annotated lexical unit for each frame,
generating one paraphrase is beneficial and generating four is even better.
With three lexical units,
the relative impact of paraphrasing begins to diminish. 
When we have one exemplar sentence per lexical unit,
adding one paraphrase is helpful, while adding four is less so. 
As the number of diverse examples increase, the impact of paraphrasing wanes.

\paragraph{Future Work} 
While we have shown that paraphrasing is beneficial for training a Frame Identification model in a low-resource setting, it is important to be aware of the limitations of paraphrastic data.  The paraphrasing generation process does not guarantee that the resulting data will be beneficial for training and evaluation since it is possible that some of the paraphrases are already well-understood by the model \citep{ribeiro-etal-2018-semantically}.
Furthermore, generated paraphrases could include lexical units that fall outside of the ontology being used, all leading to negative impact w.r.t. evaluation. Future work may investigate tactical data augmentation such as considering a filtering score proposed by \citet{ribeiro-etal-2018-semantically}
or limiting the paraphrastic data to its intersection with the FrameNet ontology.

\section{Conclusion}
We introduced a novel approach for iterative construction of semantic resources via automatic paraphrsing.  To demonstrate two possible uses of our framework, we simulated the rapid creation of a new semantic resource from a small seed corpus and generated a large-scale expansion of an existing resource. The latter experiment, run on FrameNet data, generated a lexically diverse dataset with 495,300 unique $(\texttt{Frame}, \texttt{Trigger})$ combinations annotated in context, 50x the number of such combinations originally in FrameNet, which we release to the community alongside our 36,417-instance span-alignment dataset.

\bibliography{tacl2018}
\bibliographystyle{acl_natbib}

\end{document}